\documentclass[10pt,twocolumn,letterpaper]{article}

\usepackage{cvpr}      

\usepackage[dvipsnames]{xcolor}

\definecolor{cvprblue}{rgb}{0.21,0.49,0.74}
\usepackage[pagebackref,breaklinks,colorlinks,citecolor=cvprblue]{hyperref}

\def\paperID{*****} 
\def\confName{CVPR}
\def\confYear{2024}

\title{2nd Place Solution for MOSE Track in CVPR 2024 PVUW workshop: Complex Video Object Segmentation}

\author{Zhensong Xu*, Jiangtao Yao*, Chengjing Wu, Ting Liu and Luoqi Liu\\Yao$\_$Xu$\_$MTLab, MT Lab, Meitu Inc\\
{\tt\small \{xzs1, yjt1, ethan, lt, llq\}@meitu.com}
}

\begin{document}
\maketitle
\begin{abstract}
Complex video object segmentation serves as a fundamental task for a wide range of downstream applications such as video editing and automatic data annotation. Here we present the 2nd place solution in the MOSE track of PVUW 2024. To mitigate problems caused by tiny objects, similar objects and fast movements in MOSE. We use instance segmentation to generate extra pretraining data from the valid and test set of MOSE. The segmented instances are combined with objects extracted from COCO to augment the training data and enhance semantic representation of the baseline model. Besides, motion blur is added during training to increase robustness against image blur induced by motion. Finally, we apply test time augmentation (TTA) and memory strategy to the inference stage. Our method ranked 2nd in the MOSE track of PVUW 2024, with a $\mathcal{J}$ of 0.8007, a $\mathcal{F}$ of 0.8683 and a $\mathcal{J}$\&$\mathcal{F}$ of 0.8345. 
\end{abstract}    
\section{Introduction}
\label{sec:intro}

Pixel-level Scene Understanding is one of the fundamental problems in computer vision, which aims at recognizing object classes, masks and semantics of each pixel in the given image. The pixel-level Video Understanding in the Wild Challenge (PVUW) shop challenge advances the segmentation task from images to videos, aiming at enabling challenging and practical realistic applications. The PVUW 2024 workshop challenge includes two new tracks, Complex Video Object Segmentation Track based on MOSE\cite{r9} and Motion Expression guided Video Segmentation track based on MeViS\cite{ding2023mevis}. The Complex Video Object Segmentation Track focuses on semi-supervised video object segmentation (VOS) under complex environments. As an important branch of the VOS task, semi-supervised VOS aims at tracking and segmenting agnostic objects given only the first-frame annotations, which has been widely applied in autonomous driving\cite{r1}, video editing\cite{r2}, automatic data annotation\cite{r3}, and universal video segmentation\cite{r4}.

Recent memory-based approaches have become the main stream data driven VOS methods. Memory-based approaches store past segmented frames in a memory bank, when a new query frame comes, it will read from the memory bank through cross attention, which is more robust to drifting and occlusions. Due to the advantages over other VOS methods\cite{other1,other2,other3,other4}, the memory-based paradigm has been paid much attention by the research community. As one of the most successful early attempts, Space-Time Memory network (STM)\cite{r5} stores the past frames with object masks into the memory and performs pixel-level matching between the encoded key of query frame and memory. \begin{figure}[htbp]
  \centering
  \includegraphics[width=0.98\linewidth]{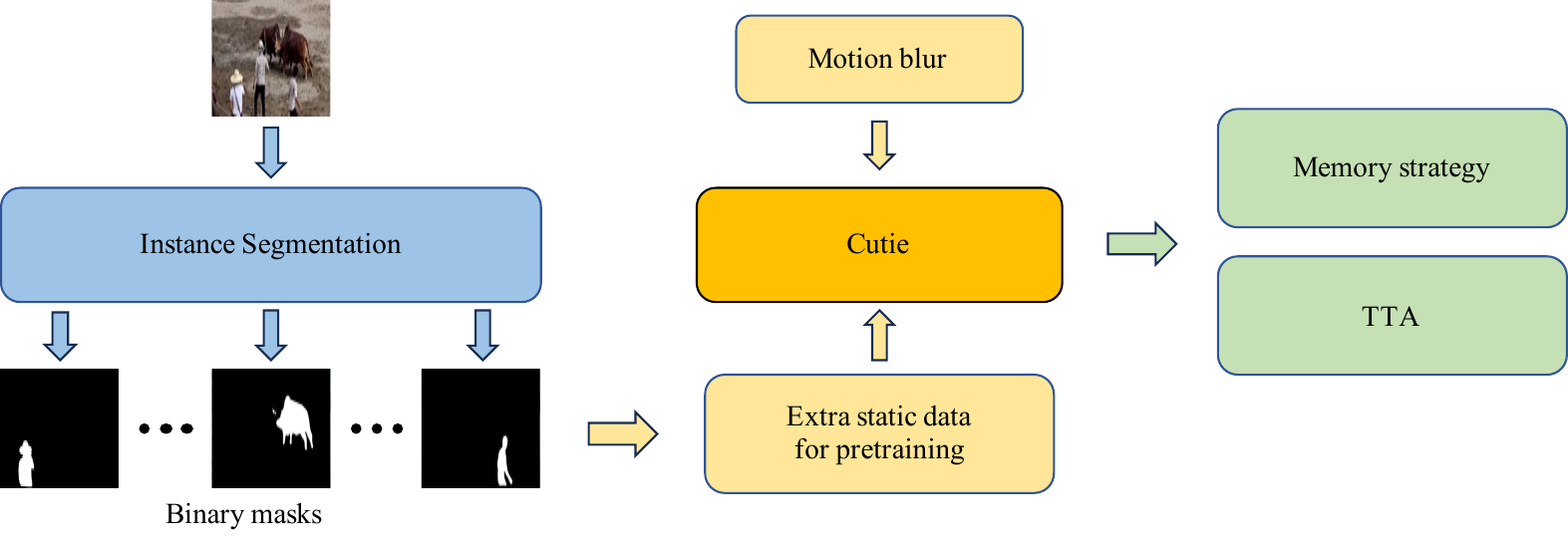}

   \caption{Overview of our method.}
   \label{fig:1}
\end{figure}
\FloatBarrier
\noindent Space-Time Correspondence Network (STCN)\cite{r6} is developed from STM, it encodes the key features from frames without masks and replace dot product with L2 similarity in the affinity for memory reading. STCN achieves better efficiency and effectiveness than STM. XMem\cite{r7} introduces three indendent memory banks: a sensory memory, a working memory and a long-term memory. The three-level memories are inspired by the Atkinson–Shiffrin memory model of Human. Xmem performs especially well on long-video datasets because of the short-term to long-term memories. To solve the mismatch problem in pixel-level matching, Cutie\cite{r8} proposes the object memory and object transformer for bidirectional information interaction. The object memory and object transformer improve robustness in challenging scenes with heavy occlusions and similarity. Due to the state-of-the-art performance of Cutie, we choose Cutie as our baseline model. 

However, the challenging nature of MOSE\cite{r9} still  
\begin{figure*}
  \centering
  \includegraphics[width=0.95\linewidth]{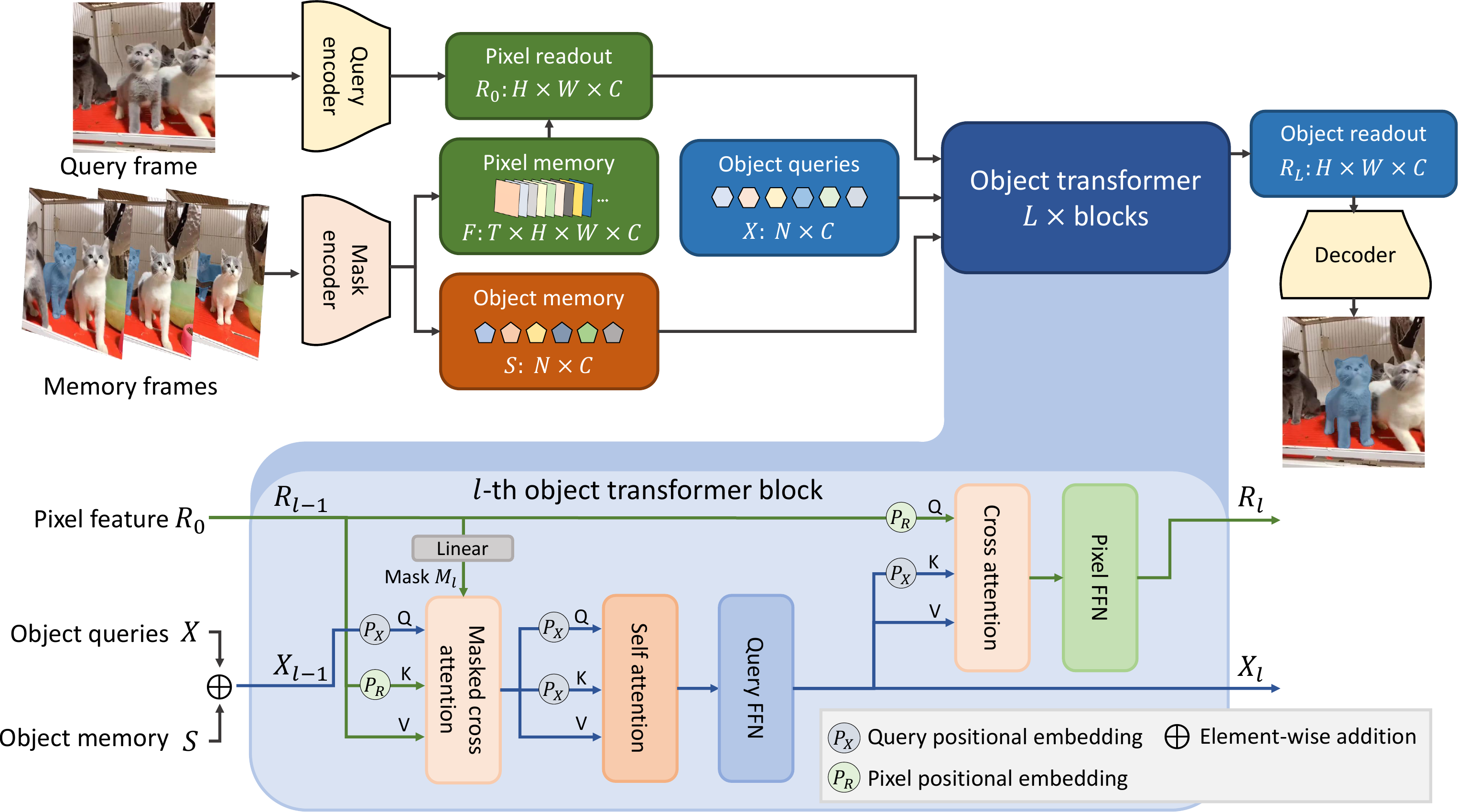}
   \caption{Architecture of Cutie\cite{r8}.}
   \label{fig:2}
\end{figure*}
\FloatBarrier
\noindent poses several obstacles to overcome. Apart from higher disappearance-reappearance rates and heavier occlusions than previous VOS datasets such as DAVIS\cite{r10} and YouTubeVOS\cite{r11}, MOSE has many tiny and similar objects, which may confuse VOS models. Besides, some videos in MOSE contain objects with fast movements, which are hard to track in consecutive frames.   

To mitigate the above issues, we combine Mask2Former\cite{r12} and motion blur as data augmentation to enhance semantic representation during the training process. We exploit images from the valid and test set of MOSE, and use Mask2Former to segment instance masks and generate  pretraining data for Cutie. We also introduce COCO\cite{r13} to further enrich the pretraining data, with the purpose of enhancing semantic learning in the early stage of training and improve the segmentation accuracy of tiny and similar objects. At inference time, we employ test time augmentation (TTA) and memory strategy to optimize the results. Our solution reaches a $\mathcal{J}$ of 0.8007, a $\mathcal{F}$ of 0.8683 and a $\mathcal{J}$\&$\mathcal{F}$ of 0.8345, which achieved the 2nd place in the MOSE track of the PVUW challenge in CVPR 2024.

\section{Method}
\label{sec:method}

As illustrated in \cref{fig:1}, our solution takes Cutie as the baseline model. Then, we use instance segmentation and motion blur to augment the training data. Finally, during the inference stage, we employ TTA and memory strategy to improve the results. Details of the solution are described as follows. 

\subsection{Baseline model}
To ensure good performance under challenges such as frequent disappearance-reappearance, heavy occlusions, small and similar objects, we introduce Cutie as the strong baseline model, as shown in \cref{fig:2}. Cutie stores a high-resolution pixel memory $F$ and a high-level object memory $S$. The pixel memory is encoded from the memory frames and corresponding segmented masks. The object memory compresses object-level features from the memory frames. When a new query frame comes, it bidirectionally interacts with the object memory in a couple of object transformer blocks. Specifically, given the feature map of the query frame, the pixel readout $R_0$ is extracted by reading from the pixel memory with a sensory memory\cite{r7}, then the pixel readout interacts with the object memory and a set of learnable object queries through bottom-up foreground-background masked cross attention. Next, the obtained high-level object query representation communicates back with the pixel readout through top-down cross attention. The output pixel readout $R_l$ and object queries $X_l$ are sent to the next object transformer block. The final pixel readout will be combined with multi-scale features passed from skip connections for computing the output mask in the decoder. Cutie enriches pixel features with object-level semantics in a bidirectional fashion, hence is more robust to distractions 

\begin{figure*}
    \centering
    \begin{tabular}{c@{\hspace{1pt}}c@{\hspace{1pt}}c@{\hspace{1pt}}c}
    \includegraphics[width=0.32\linewidth, height=0.20\linewidth]{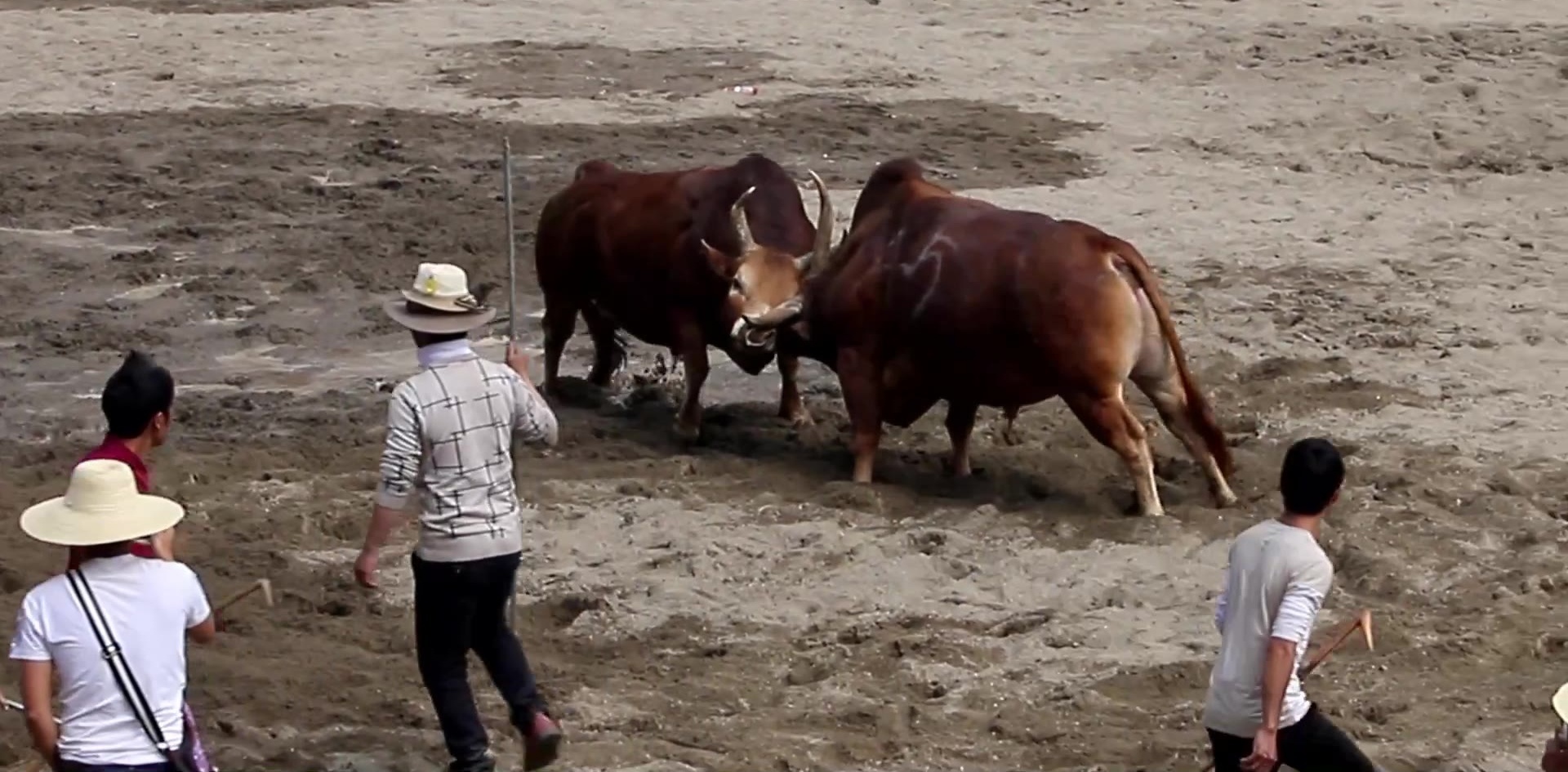} & 
        \includegraphics[width=0.32\linewidth, height=0.20\linewidth]{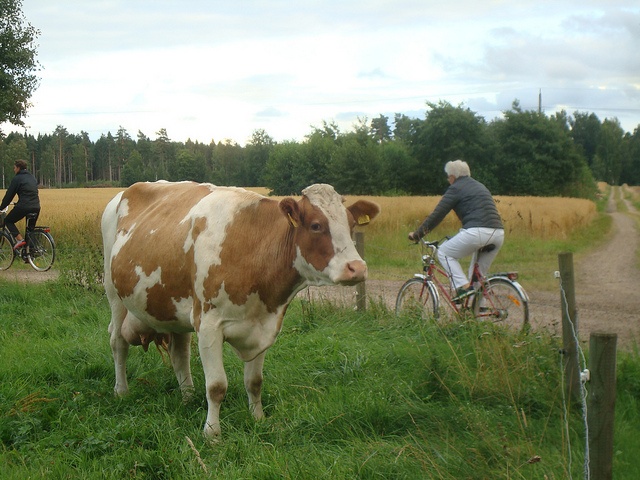} & 
        \includegraphics[width=0.32\linewidth, height=0.20\linewidth]{fig/Fig_3/bike.jpg} \\
        \includegraphics[width=0.32\linewidth, height=0.20\linewidth]{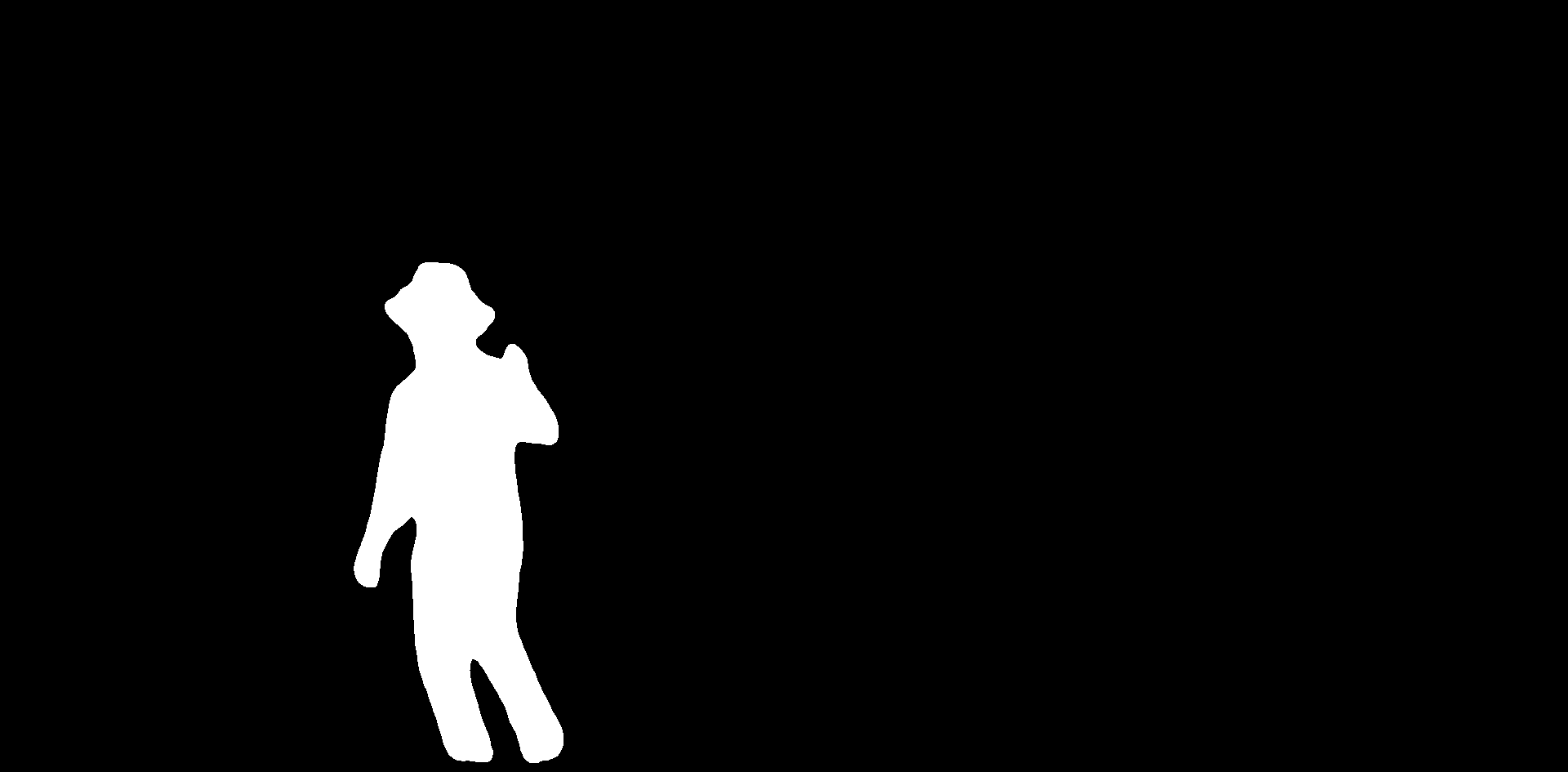} & 
        \includegraphics[width=0.32\linewidth, height=0.20\linewidth]{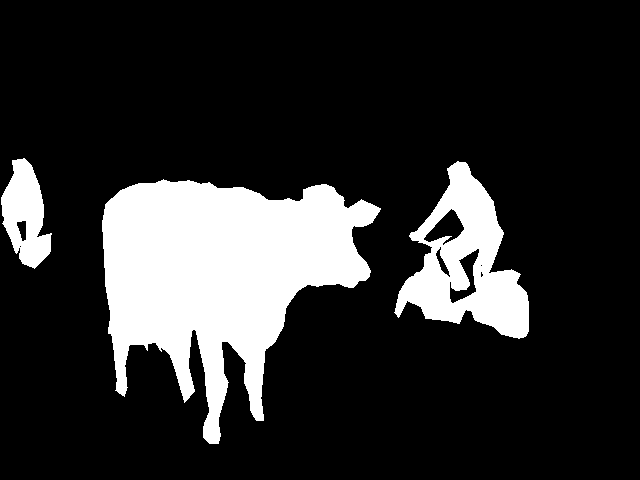} & 
        \includegraphics[width=0.32\linewidth, height=0.20\linewidth]{fig/Fig_3/bike_blur.jpg} \\
     \end{tabular}
    \caption{Examples of generated pretraining data and motion blur. Left: binary mask generated from the valid set and test set of MOSE. Middle: binary mask generated from COCO, the masks of different classes are merged into one mask. Right: example of motion blur in the horizontal direction.}
    \label{fig:3}
\end{figure*}
\FloatBarrier 
\noindent such as occlusion and disappearance. 

\subsection{Data augmentation}
Like most state-of-the-art VOS methods, Cutie also adopts a two-stage training paradigm. The first stage pretraining uses short video sequences generated from static images. Then main training is performed using VOS datasets in the second stage. However, the original Cutie fails to perform well when similar objects move in close proximity or suffers from serious motion blur. 

To solve the above problems, we conduct data augmentation to enhance the training of Cutie. First, we employ the universal image segmentation model Mask2Former to segment instance targets from the valid set and test set of MOSE. As shown in the left column of \cref{fig:3}, the segmented small objects represent typical object appearances in MOSE, which is helpful for learning the semantics of diverse objects in advance. Meanwhile, as shown in the middle column of \cref{fig:3}, we convert the instance annotations of COCO into independent binary masks. Here we select object classes such as human, animal and vehicle that frequently occur in MOSE to reduce  discrepancy between two data distributions. The acquired data is used as extra pretraining data to enable more robust semantics and improve discrimination ability against diverse objects of MOSE. Second, with the observation that motion blur is a significant challenge, we add motion blur with random kernel sizes and angles to both the pretraining and main training stages. An example of motion blur is shown in the right column of \cref{fig:3}. The proposed data augmentation aims at training towards better robustness and generalization. 

\subsection{Inference time operations}
\paragraph{TTA.} We use two kinds of TTA: flipping and multi-scale data enhancement. We only conduct horizontal flipping since experiments show flipping in other directions is detrimental to performance. In addition, we inference results on the test set under three maximum shorter side resolutions: 600p, 720p and 800p. The multi-scale results are then averaged to get the final result.\\
\textbf{Memory strategy.} We find in experiments that larger memory banks and shorter memory intervals lead to better performance. Therefore, we adjust the maximum memory frames $T_{\max}$ to 18 and the memory interval to 1. 

\section{experiment}
\label{sec:experiment}

\subsection{Implementation details}
\textbf{Data.} ECSSD\cite{r14}, DUTS\cite{r15}, FSS-1000\cite{r16}, HRSOD\cite{r17} and BIG\cite{r18} are used as image segmentatio datasets for pretraining. Besides, we generate 66823 image-mask pairs from the valid and test set of MOSE  using Mask2Former and 89490 image-mask pairs from COCO, and add them into the data for pretraining. For main training, we mix the training sets of DAVIS-2017\cite{r10}, YouTubeVOS-2019\cite{r11}, BURST\cite{r3}, OVIS\cite{r19} and MOSE\cite{r9}. \\
\textbf{Training.} The parameter is updated using AdamW\cite{r20} with a learning rate of 0.0001, a batch size of 16, and a weight decay of 0.001. Pretraining is carried out for 80K iterations with a crop size of 384×384 and no learning rate decay. Main training is carried out for 175K iterations, with 

\begin{figure*}
  \centering
  \includegraphics[width=0.95\linewidth]{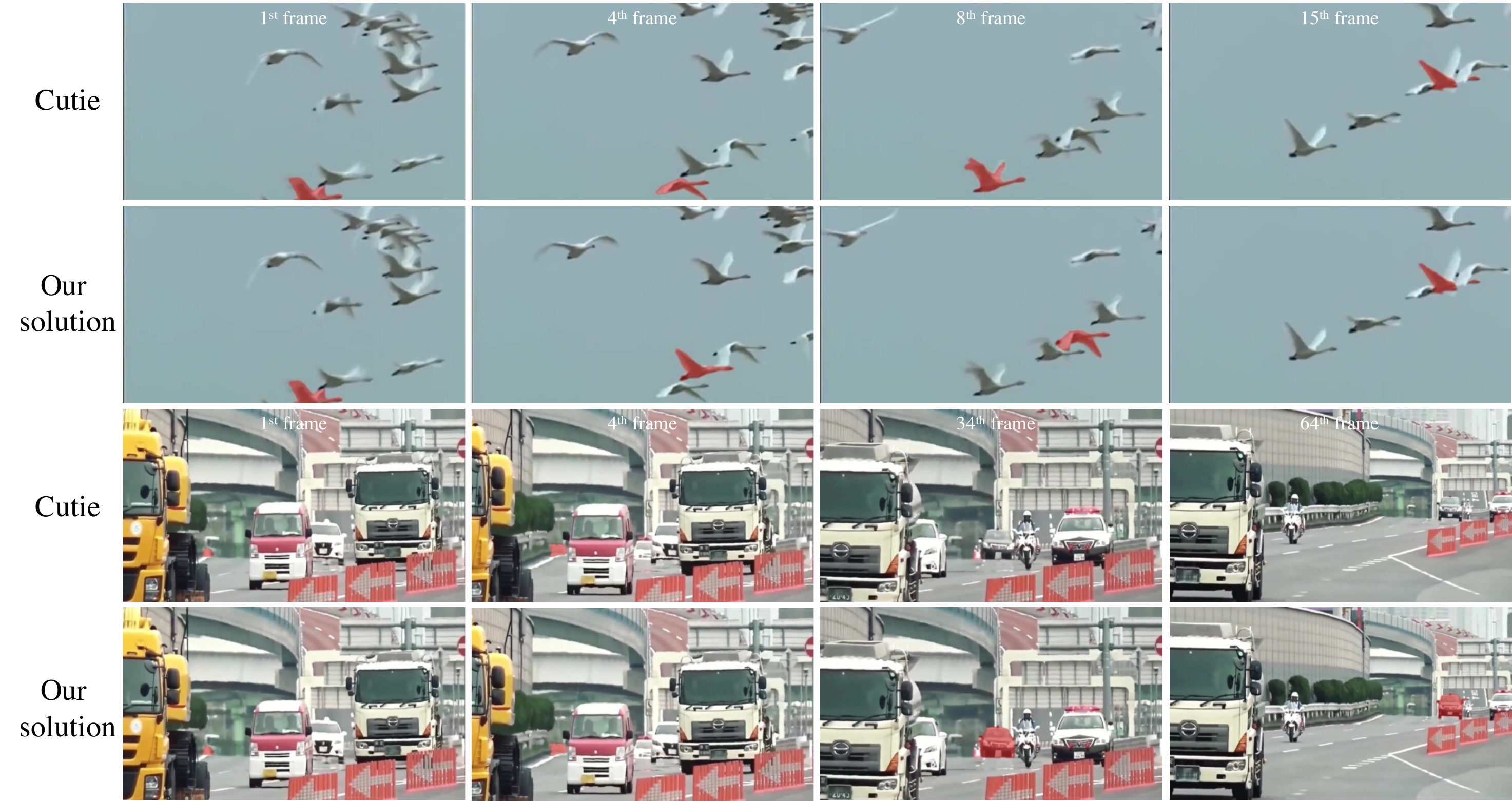}
   \caption{Qualitative results on the test set of MOSE.}
   \label{fig:4}
\end{figure*}
\FloatBarrier

\begin{table}[htbp]
\setlength{\tabcolsep}{4mm}
\centering
\begin{tabular}{cccc}
\toprule
	Method & $\mathcal{J}$ & $\mathcal{F}$ & $\mathcal{J}$\&$\mathcal{F}$\\
	\hline
    PCL$\_$MDS & 0.8101 &	0.8789 & 0.8445 \\
    \textcolor{blue}{Yao$\_$Xu$\_$MTLab} & \textcolor{blue}{0.8007} & \textcolor{blue}{0.8683} & \textcolor{blue}{0.8345} \\
    ISS & 0.7879 & 0.8559 &	0.8219 \\ 
    xsong2023 & 0.7873 & 0.8544 & 0.8208 \\
    yangdonghan50 & 0.7799 & 0.8480 & 0.8139 \\

\bottomrule
 \end{tabular}
\caption{Leaderboard of the 1st MOSE challenge. Our results are marked in blue.}
 \label{tab:1}
\end{table}
\FloatBarrier

\noindent a crop size of 480×480 and the learning rate reduced by 10 times after 140K and 160K iterations. 

\subsection{Results in the 1st MOSE challenge}
Our proposed method ranked the 2nd place in the 1st MOSE challenge. The leaderboard is displayed in \cref{tab:1}. Our method achieved a $\mathcal{J}$ of 0.8007, a $\mathcal{F}$ of 0.8683 and a $\mathcal{J}$\&$\mathcal{F}$ of 0.8345. 

\subsection{Ablation study}
We conduct an ablation study to verify the effectiveness of different components in our method. Specifically, we take the original Cutie as the baseline, then we incorporate the data augmentation, TTA and memory strategy into the baseline and design two ablation variants. From the quantitative results in \cref{tab:2}, data augmentation through instance segmentation and motion blur improves the $\mathcal{J}$\&$\mathcal{F}$ for about 0.0184. Test time augmentation and memory strategy bring the most significant improvement, with about 0.03 increase in all three metrics. As shown by the qualitative results in \cref{fig:4}, our solution performs better when segmenting tiny and similar objects with movements.

\begin{table}[htbp]
\setlength{\tabcolsep}{2.5mm}
\centering
\begin{tabular}{cccc}
\toprule
	Method & $\mathcal{J}$ & $\mathcal{F}$ & $\mathcal{J}$\&$\mathcal{F}$\\
	\hline
    Baseline & 0.7509 &	0.8206 & 0.7857 \\
    Baseline+DA & 0.7713 & 0.8373 & 0.8043 \\ 
    Baseline+DA+TTA+MS & 0.8007 & 0.8683 & 0.8345 \\
\bottomrule
 \end{tabular}
\caption{Quantitative results of the ablation study. DA indicates data augmentation, TTA indicates test time augmentation and MS indicates memory strategy.}
 \label{tab:2}
\end{table}
\FloatBarrier

\section{Conclusion}
In this paper, we propose a method for complex video object segmentation. Specifically, we take Cutie as the baseline model, and conduct data augmentation to enhance feature learning through Mask2Former and motion blur. TTA and memory strategy are employed in the inference stage to improve the segmentation results. Our method achieved the 2nd place on the MOSE track of the PVUW Challenge 2024 with 0.8345 $\mathcal{J}$\&$\mathcal{F}$.
{
    \small
    \bibliographystyle{unsrt}
    \bibliography{main}
}


\end{document}